\documentclass[12pt,journal,compsoc,onecolumn]{IEEEtran}

\usepackage{chapterbib}
\usepackage[sectionbib,numbers]{natbib}
\usepackage{booktabs}
\usepackage{multirow}
\usepackage{tabularx}
\usepackage{amsmath}
\usepackage{amssymb}
\usepackage{cleveref}
\usepackage{makecell}
\usepackage{caption}
\usepackage{breakcites}
\usepackage{graphicx}
\usepackage{amsfonts}
\usepackage{bm}
\usepackage{amsthm}
\usepackage{float}
\usepackage{pifont}
\newcommand{\cmark}{\ding{51}}
\newcommand{\xmark}{\ding{55}}
\usepackage{color}
\usepackage{soul}

\renewcommand{\normalsize}{\fontsize{11}{12}\selectfont}

\begin{document}

\title{Deep Learning for Image Search and Retrieval in Large Remote Sensing Archives}
\author{\IEEEauthorblockN{Gencer~Sumbul,
        Jian~Kang,
        and Beg\"um Demir\\}\vspace{0.5cm}%
\IEEEauthorblockA{\normalsize Faculty of Electrical Engineering and Computer Science, Technische Universit\"at Berlin, Germany}%
\IEEEcompsocitemizethanks{\IEEEcompsocthanksitem This work is funded by the European Research Council (ERC) through the ERC-2017-STG BigEarth Project under Grant 759764. (Corresponding Author: Beg\"um Demir (email: demir@tu-berlin.de)).}
}


\IEEEtitleabstractindextext{
\begin{abstract}
\normalsize
This chapter presents recent advances in content based image search and retrieval (CBIR) systems in remote sensing (RS) for fast and accurate information discovery from massive data archives. Initially, we analyze the limitations of the traditional CBIR systems that rely on the hand-crafted RS image descriptors. Then, we focus our attention on the advances in RS CBIR systems for which deep learning (DL) models are at the forefront. In particular, we present the theoretical properties of the most recent DL based CBIR systems for the characterization of the complex semantic content of RS images. After discussing their strengths and limitations, we present the deep hashing based CBIR systems that have high time-efficient search capability within huge data archives. Finally, the most promising research directions in RS CBIR are discussed. 
\end{abstract}}

\maketitle
\IEEEdisplaynontitleabstractindextext

\section{Introduction}
\label{chp:introduction}

\begin{figure}[t]
\centering
\includegraphics[width=0.75\textwidth]{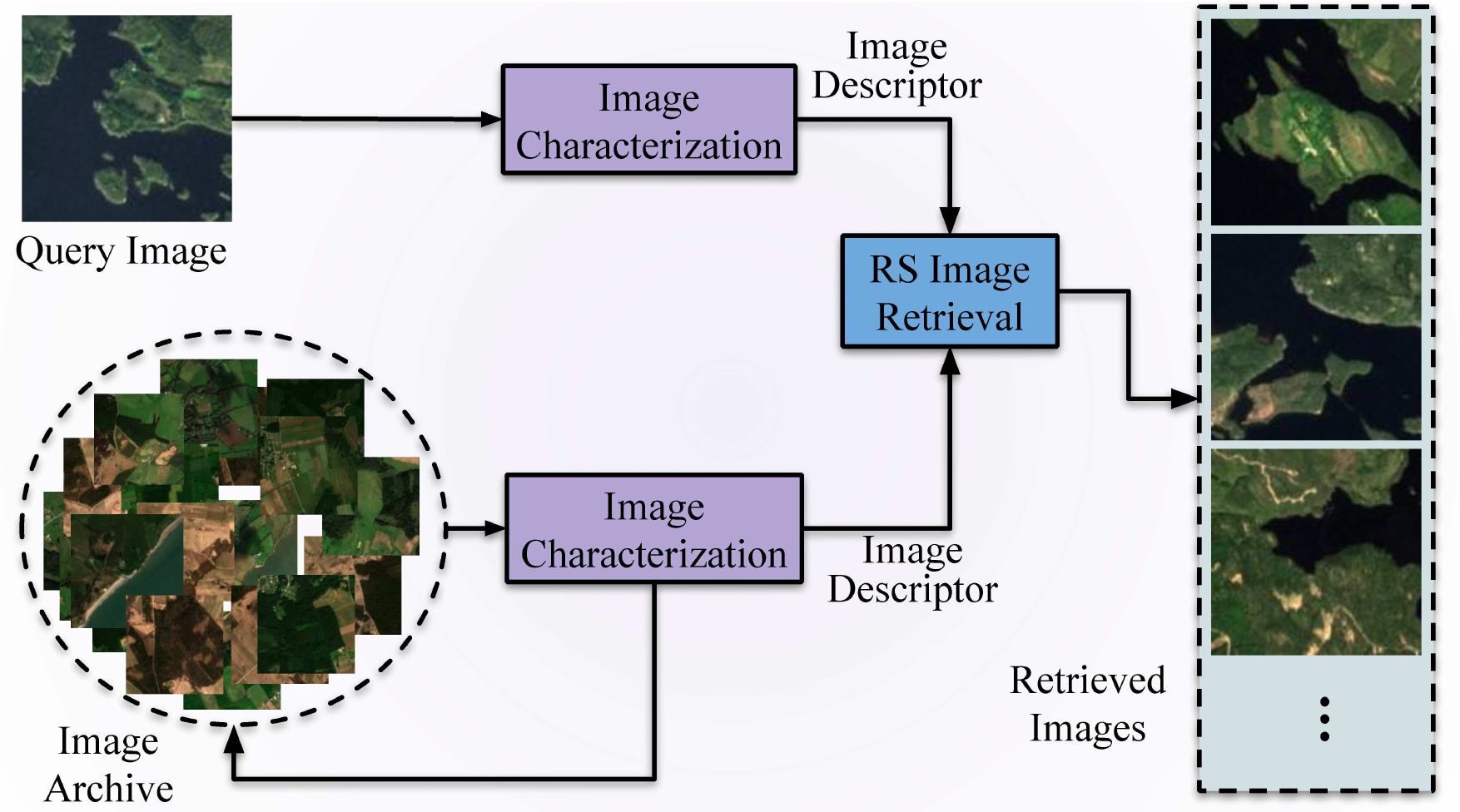}
\caption{General block scheme of a RS CBIR system.\label{fig1}}
\end{figure}

With the unprecedented advances in the satellite technology, recent years have witnessed a significant increase in the volume of remote sensing (RS) image archives. Thus, the development of efficient and accurate content based image retrieval (CBIR) systems in massive archives of RS images is a growing research interest in RS. CBIR aims to search for RS images of the similar information content within a large archive with respect to a query image. To this end, CBIR systems are defined based on two main steps: i) image description step (which characterizes the spatial and spectral information content of RS images); and ii) image retrieval step (which evaluates the similarity among the considered descriptors and then retrieve images similar to a query image in the order of similarity). A general block scheme of a CBIR system is shown in \Cref{fig1}.

Traditional CBIR systems extract and exploit hand-crafted features to describe the content of RS images. As an example, bag-of-visual-words representations of the local invariant features extracted by the scale invariant feature transform (SIFT) are introduced in \citep{Yang:2013}. In \citep{Aptoula:2014}, a bag-of-morphological-words representation of the local morphological texture features (descriptors) is proposed in the context of CBIR. Local Binary Patterns (LBPs), which represent the relationship of each pattern (i.e., pixel) in a given image with its neighbors located on a circle around that pixel, are found very efficient in RS. In \citep{Tekeste:20118}, a comparative study that analyzes and compares different LBPs in RS CBIR problems is presented. To define the spectral information content of high dimensional RS images the bag-of-spectral-values descriptors are presented in \citep{Dai:2018}. Graph-based image representations, where the nodes describe the image region properties and the edges represent the spatial relationships among the regions, are presented in \citep{Li:2007}, \citep{Chaudhuri:2016}, \citep{Chaudhuri:2018}. Hashing methods that embed high-dimensional image features into a low-dimensional Hamming (binary) space by a set of hash functions are found very effective in RS \citep{Demir:2016}, \citep{Li:2017}, \citep{Reato:2019}. By this way, the images are represented by binary hash codes that can significantly reduce the amount of memory required for storing the RS images with respect to the other descriptors. Hashing methods differ from each other on how the hash functions are generated. As an example, in \citep{Demir:2016}, \citep{Reato:2019} kernel-based hashing methods that define hash functions in the kernel space are presented, whereas a partial randomness hashing method that defines the hash functions based on a weight matrix defined using labeled images is introduced in \citep{Li:2017}. More details on hashing for RS CBIR problems are given in Section 7.4.

Once image descriptors are obtained, one can use the k-nearest neighbor (\textit{k}-NN) algorithm, which computes the similarity between the query image and all archive images to find the k most similar images to the query. If the images are represented by graphs, graph matching techniques can be used. As an example, in \citep{Chaudhuri:2016} an inexact graph matching approach, which is based on the sub-graph isomorphism and spectral embedding algorithms, is presented. If the images are represented by binary hash codes, image retrieval can be achieved by calculating the Hamming distances with simple bit-wise XOR operations that allow time-efficient search capability \citep{Demir:2016}. However, these unsupervised systems do not always result in satisfactory query responses due to the semantic gap, which is occurred among the low-level features and the high-level semantic content of RS images \citep{Demir:2015}. To overcome this problem and improve the performance of CBIR systems, semi-supervised and fully supervised systems, which require user feedback in terms of RS image annotations, are introduced \citep{Demir:2015}. Most of these systems depend on the availability of training images, each of which is annotated with a single broad category label that is associated to the most significant content of the image. However, RS images typically contain multiple classes and thus can simultaneously be associated with different class labels. Thus, CBIR methods that properly exploit training images annotated by multi-labels are recently found very promising in RS. As an example, in \citep{Dai:2018} a CBIR system that exploits a measure of label likelihood based on a sparse reconstruction-based classifier is presented in the framework of multi-label RS CBIR problems. Semi-supervised CBIR systems based on graph matching algorithms are proposed in \citep{Wang:2016}, \citep{Chaudhuri:2018}. In detail, in \citep{Wang:2016} a three-layer framework in the context graph-based learning is proposed for query expansion and fusion of global and local features by using the label information of query images. In \citep{Chaudhuri:2018} a correlated label propagation algorithm, which operates on a neighborhood graph for automatic labeling of images by using a small number of training images, is proposed. 

The above-mentioned CBIR systems rely on shallow learning architectures and hand-crafted features. Thus, they can not simultaneously optimize feature learning and image retrieval, resulting in limited capability to represent the high-level semantic content of RS images. This issue leads to inaccurate search and retrieval performance in practice. Recent advances in deep neural networks (DNNs) have triggered substantial performance gain for image retrieval due to their high capability to encode higher level semantics present in RS images. Differently from conventional CBIR systems, deep learning (DL) based CBIR systems learn image descriptors in such a way that feature representations are optimized during the image retrieval process. In order words, DNNs eliminate the need for human effort to design discriminative and descriptive image descriptors for the retrieval problems. Most of the existing RS CBIR systems based on DNNs attempt to improve image retrieval performance by: 1) learning discriminative image descriptors; and 2) achieving scalable image search and retrieval. The aim of this chapter is to present different DNNs proposed in the literature for the retrieval of RS images. The rest of this chapter is organized as follows. Section 7.3 reviews the DNNs proposed in the literature for the description of the complex information content of RS images in the framework of CBIR. Section 7.4 presents the recent progress on the scalable CBIR systems defined based on DNNs in RS. Finally, Section 7.5 draws the conclusion of this chapter.

\section{Deep Learning for RS CBIR}
\label{chp:first}
 
\begin{figure}[t]
\centering
\includegraphics[width=0.8\textwidth]{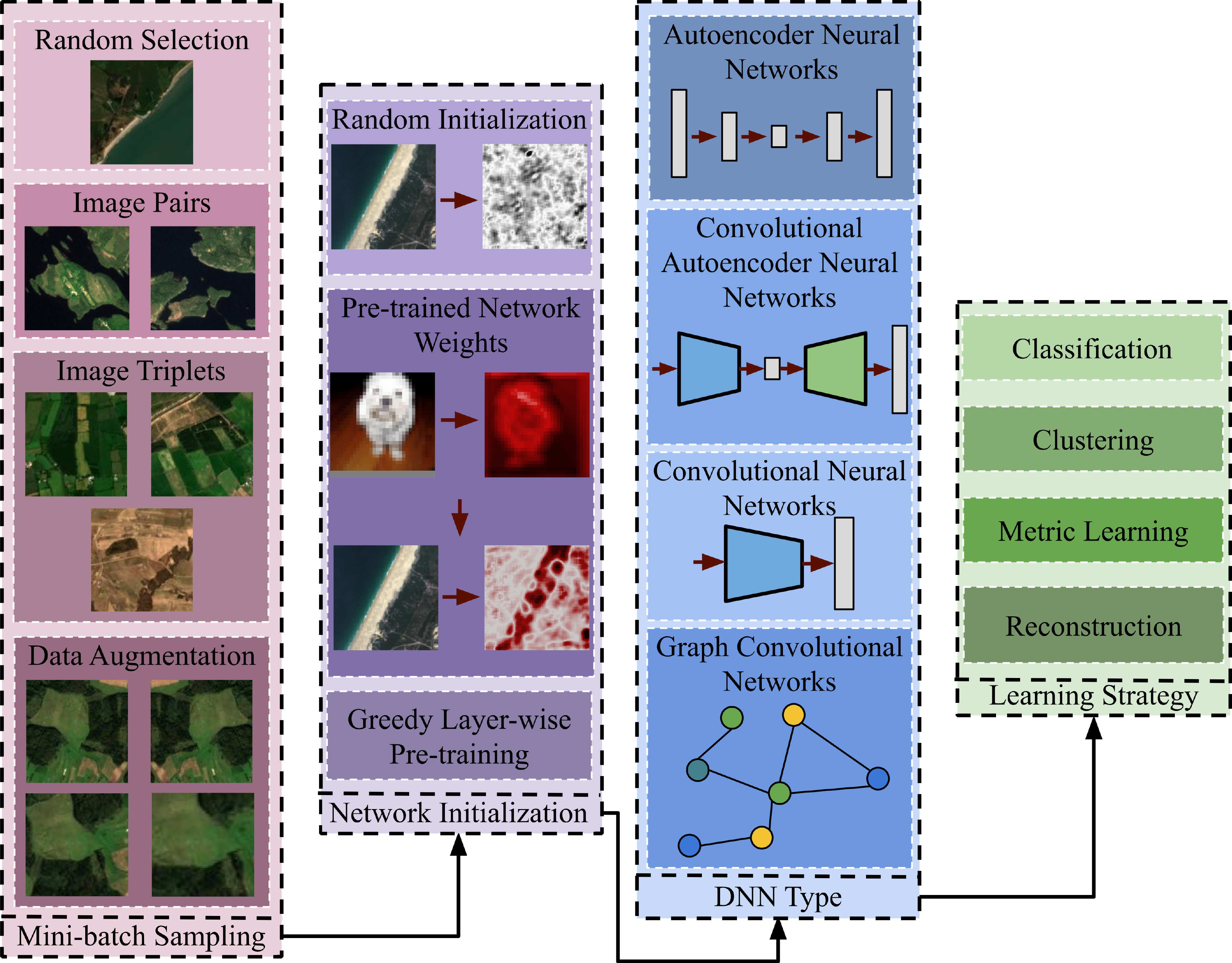}
\caption{Different strategies considered within the DL based RS CBIR systems.\label{deep_CBIR}}
\end{figure}

 The DL based CBIR systems in RS differ from each other in terms of: i) the strategies considered for the mini-batch sampling; ii) the approaches used for the initialization of the parameters of the considered DNN model; iii) the type of the considered DNN; and iv) the strategies used for image representation learning. \Cref{deep_CBIR} illustrates the main approaches utilized in DL based CBIR systems in RS. In detail, a set of training images is initially selected from the considered archive to train a DNN. Then, the selected training images are divided into mini-batches and fed into the considered DNN. After initializing the model parameters of the network, the training phase is conducted with an iterative estimation of the model parameters based on a loss function. The loss function is selected on the basis of the characteristics of the considered learning strategy. 

During the last years, several DL based CBIR systems that consider different strategies for the above-mentioned factors are presented. As an example, in \citep{Zhou:2015} an unsupervised feature learning framework that learns image descriptors from a set of unlabeled RS images based on an autoencoder (AE) is introduced. After random selection of mini-batches and initialization of the model parameters, SIFT based image descriptors are encoded into sparse descriptors by learning the reconstruction of the descriptors. The learning strategy relies on minimization of a reconstruction loss function between the SIFT descriptors and the reconstructed image descriptors in the framework of the AE. A CBIR system that applies a multiple feature representation learning and a collaborative affinity metric fusion is presented in \citep{Li:2016}. This system randomly selects RS images for mini-batches and initializes the model parameters of a Convolutional Neural Network (CNN). Then, it employs the CNN for k-means clustering (instead of classification). To this end, a reconstruction loss function is utilized to minimize the error induced between the CNN results and the cluster assignments. Collaborative affinity metric fusion is employed to incorporate the traditional image descriptors (e.g., SIFT, LBP) with those extracted from different layers of the CNN. A CBIR system with deep bag-of-words is proposed in \citep{Xu:2018}. This system employs a convolutional autoencoder (CAE) for extracting image descriptors in an unsupervised manner. The method first encodes the local areas of randomly selected RS images into a descriptor space and then decodes from descriptors to image space. Since encoding and decoding steps are based on convolutional layers, a reconstruction loss function is directly applied to reduce the error between the input and constructed local areas for the unsupervised reconstruction based learning. Since this system operates on local areas of the images, bag-of-words approach with k-means clustering is applied to define the global image descriptor from local areas. Although this system has the same learning strategy as \citep{Zhou:2015}, its advantages are two-fold compared to \citep{Zhou:2015}. First, model parameters are initialized with greedy layer-wise pre-training that allows more effective learning procedure with respect to the random initialization approach. Second, the CAE model has better capability to characterize the semantic content of images since it considers the neighborhood relationship through the convolution operations. 

Reconstruction based unsupervised learning of RS image descriptors is found effective particularly when annotated training images are not existing. However, minimizing a reconstruction loss function on a small amount of unannotated images with a shallow neural network limits the accurate description of the high-level information content of RS images. This problem can be addressed by supervised DL-based CBIR systems that require a training set that consists of a high number of annotated images to learn effective models with several different parameters. The amount and the quality of the training images determine the success of the supervised DL models. However, annotating RS images at large-scale is time consuming, complex, and costly in operational applications. To overcome this problem, a common approach is to exploit DL models with proven architectures (such as ResNet or VGG), which are pre-trained on publicly available general purpose computer vision (CV) datasets (e.g., ImageNet). The existing models are then fine-tuned on a small set of annotated RS images to calibrate the final layers (this is known as transfer learning). As an example, in \citep{Hu:2016} model parameters of a CNN are initialized with the parameters of a CNN model that is pre-trained on ImageNet. In this work, both initial training and fine-tuning are applied in the framework of the classification problems. To this end, the cross-entropy loss function is utilized to reduce the class prediction errors. Image descriptors learned with the cross-entropy loss function encode the class discrimination instead of similarities among images. Thus, it can limit the performance of CBIR systems. A data augmentation technique for mini-batch sampling is utilized in \citep{Hu:2016} to improve the effectiveness of the image descriptors. To this end, different scales of RS images in a mini-batch are fed into the CNN. Then, the obtained descriptors are aggregated by using different pooling strategies to characterize the final image descriptors. A low-dimensional convolutional neural network (LDCNN) proposed in \citep{Zhou:2017} also utilizes parameters of a pre-trained DL model to initialize the network parameters. However, it randomly selects RS images for the definition of mini-batches and adopts a classification based learning strategy with the cross-entropy loss function. This system combines convolutional layers with cross channel parametric pooling and global average pooling to characterize low dimensional descriptors. Since fully connected (FC) layers are replaced with pooling layers, LDCNN significantly decreases the total number of model parameters required to be estimated during the training phase. This leads to significantly reduced computational complexity and also reduced risk of over-fitting (which can occur in the case of training CNNs with a small amount of training images). A CBIR system based on a CNN with weighted distance is introduced in \citep{Ye:2018}. Similar to \citep{Hu:2016} and \citep{Zhou:2017}, this system also applies fine-tuning on a state-of-the-art CNN model pre-trained on ImageNet. In addition, it enhances the conventional distance metrics used for image retrieval by weighting the distance between a query image and the archive images based on their class probabilities obtained by a CNN. An enhanced interactive RS CBIR system, which extracts the preliminary RS image descriptors based on the LDCNN by utilizing the same mini-batch sampling, network initialization and learning strategy (based on the cross-entropy loss function), is introduced in \citep{Boualleg:2018}. Labeled training images are utilized to obtain the preliminary image descriptors. Then, a relevance feedback scheme is applied to further improve the effectiveness of the image descriptors by considering the user feedbacks on the automatically retrieved images. The use of aggregated deep local features for RS image retrieval is proposed in \citep{Imbriaco:2019}. To this end, the VLAD representation of local convolutional descriptors from multiplicative and additive attention mechanisms are considered to characterize the descriptors of the most relevant regions of the RS images. This is achieved based on three steps. In the first step, similar to \citep{Ye:2018} and \citep{Zhou:2017}, the system operates on randomly selected RS images and applies fine-tuning to a state-of-the art CNN model while relying on a classification based learning strategy with the cross-entropy loss function. In the second step, additive and multiplicative attention mechanisms are integrated into the convolutional layers of the CNN and thus are retrained to learn their parameters. Then, local descriptors are characterized based on the attention scores of the resized RS images at different scales (which is achieved based on data augmentation). In the last step, the system transforms VLAD representations with Memory Vector (MV) construction (which produces the expanded query descriptor) to make the CBIR system sensitive to the selected query images. In this system, the query expansion strategy is applied after obtaining all the local descriptors. This query-sensitive CBIR approach further improves the discrimination capability of image descriptors, since it adapts the overall learning procedure of DNNs based on the selected queries. Thus, it has a huge potential for RS CBIR problems. 

Most of the above-mentioned DL based supervised CBIR systems learn an image feature space directly optimized for a classification task by considering entropy-based loss functions. Thus, the image descriptors are designed to discriminate the pre-defined classes by taking into account the class based similarities rather than the image based similarities during the training stage of the DL models. The absence of positive and negative images with respect to the selected query image during the training phase can lead to a poor CBIR performance. To overcome this limitation, metric learning is recently introduced in RS to take into account image similarities within DNNs. Accordingly, a Siamese graph convolutional network is introduced in \citep{Chaudhuri:2019} to model the weighted region adjacency graph (RAG) based image descriptors by a metric learning strategy. To this end, mini-batches are first constructed to include either similar or dissimilar RS images (Siamese pairs). If a pair of images belongs to the same class, they are assumed as similar images, and vice versa. Then, RAGs are fed into two graph convolutional networks with shared parameters to model image similarities with the contrastive loss function. Due to the considered metric learning strategy (which is guided by the contrastive loss function) the distance between the descriptors of similar images is decreased, while that between dissimilar images is increased. The contrastive loss function only considers the similarity estimated among image pairs, i.e., similarities among multiple images are not evaluated, which can limit the success of similarity learning for CBIR problems. 

To address this limitation, a triplet deep metric learning network (TDMLN) is proposed in \citep{Cao:2019}. TDMLN employs three CNNs with shared model parameters for similarity learning through image triplets in the content of metric learning. Model parameters of the TDMLN are initialized with a state-of-the-art CNN model pre-trained on ImageNet. For the mini-batch sampling, TDMLN considers an anchor image together with a similar (i.e., positive) image and a dissimilar (i.e., negative) image to the anchor image at a time. Image triplets are constructed based on the annotated training images \citep{Chaudhuri:2019}. While anchor and positive images belong to the same class, the negative image is associated to a different class. Then, similarity learning of the triplets is achieved based on the triplet loss function. By the use of triplet loss function, the distance estimated between the anchor and positive images in the descriptor (i.e., feature) space is minimized, whereas that computed between the anchor and negative images is separated by a certain margin.
\begin{figure}
    \centering
	\includegraphics[width=0.6\textwidth]{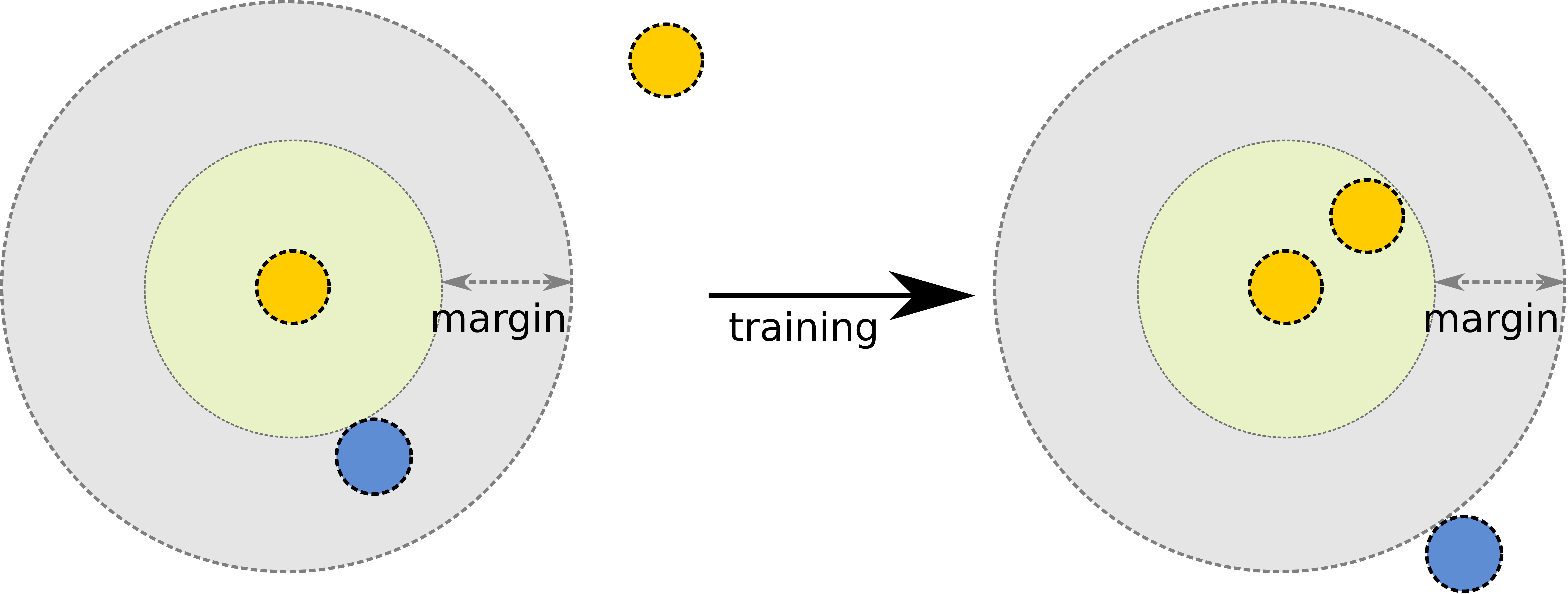}
	\caption{The intuition behind the triplet loss function: after training, a positive sample is moved closer to the anchor sample than the negative samples of the other classes.}
	\label{fig:triplet_loss}
\end{figure}
\Cref{fig:triplet_loss} illustrates intuition behind the triplet loss function. Metric learning guided by the triplet loss function learns similarity based on the image triplets and thus provides highly discriminative image descriptors in the framework of CBIR. However, how to define and select image triplets is still an open question. Current methods rely on the image-level annotations based on the land-cover land-use class labels, which do not directly represent the similarity of RS images. Thus, metric learning based CBIR systems need further improvements to characterize retrieval specific image descriptors. 
\begin{table}[t]
\footnotesize
\setlength{\tabcolsep}{5pt}
\centering
\caption{Main characteristics of the DL based CBIR systems in RS.\label{tab2}}{%
\begin{tabular}{@{}clllll@{}}
\toprule
\parbox{1.35701cm}{\textbf{Reference}} & 
\parbox{2.061cm}{\textbf{Mini-batch\\Sampling}} & 
\parbox{2.075cm}{\textbf{Network\\Initialization}} & 
\parbox{2.093cm}{\textbf{DNN Type}} & 
\parbox{2.081cm}{\textbf{Learning\\Strategy}} & 
\parbox{2cm}{\textbf{Loss\\Function}} \\ \midrule
\parbox{1.35701cm}{\citep{Zhou:2015}} & \parbox{2.061cm}{Random\\selection}  & \parbox{2.075cm}{Random\\initialization} & \parbox{2.093cm}{Auto-encoder} & \parbox{2.081cm}{Reconstruction\\(unsupervised)} & \parbox{2cm}{Reconstruction} \\ \midrule
\parbox{1.35701cm}{\citep{Hu:2016}} & \parbox{2.061cm}{Data\\augmentation} & \parbox{2.075cm}{Pre-trained\\network weights} & \parbox{2.093cm}{Convolutional\\neural network} & \parbox{2.081cm}{Classification\\(supervised)} & \parbox{2cm}{Cross-entropy} \\ \midrule
\parbox{1.35701cm}{\citep{Li:2016}} & \parbox{2.061cm}{Random\\selection}  & \parbox{2.075cm}{Random\\initialization} & \parbox{2.093cm}{Convolutional\\neural network} & \parbox{2.081cm}{Clustering\\ (unsupervised)} & \parbox{2cm}{Reconstruction} \\ \midrule
\parbox{1.35701cm}{\citep{Zhou:2017}} &
\parbox{2.061cm}{Random\\selection} & \parbox{2.075cm}{Pre-trained\\network weights} & \parbox{2.093cm}{Convolutional\\neural network} & \parbox{2.081cm}{Classification\\(supervised)} & \parbox{2cm}{Cross-entropy} \\ \midrule
\parbox{1.35701cm}{\citep{Ye:2018}} &
\parbox{2.061cm}{Random\\selection} & \parbox{2.075cm}{Pre-trained\\network weights} & \parbox{2.093cm}{Convolutional\\neural network} & \parbox{2.081cm}{Classification\\(supervised)} & \parbox{2cm}{Cross-entropy} \\ \midrule
\parbox{1.35701cm}{\citep{Xu:2018}} & \parbox{2.061cm}{Random\\selection}  & \parbox{2.075cm}{Greedy\\layer-wise\\pre-training } & \parbox{2.093cm}{Convolutional\\auto-encoder} & \parbox{2.081cm}{Reconstruction\\(unsupervised)} & \parbox{2cm}{Reconstruction} \\ \midrule
\parbox{1.35701cm}{\citep{Boualleg:2018}} & 
\parbox{2.061cm}{Random\\Selection} & \parbox{2.075cm}{Pre-trained\\network weights} & \parbox{2.093cm}{Convolutional\\neural network} & \parbox{2.081cm}{Classification\\(supervised)} & \parbox{2cm}{Cross-entropy} \\ \midrule
\parbox{1.35701cm}{\citep{Imbriaco:2019}} & \parbox{2.061cm}{Random\\selection\\Data\\augmentation} & \parbox{2.075cm}{Pre-trained\\network weights} & \parbox{2.093cm}{Convolutional\\neural network} & \parbox{2.081cm}{Classification\\(supervised)} & \parbox{2cm}{Cross-entropy} \\\midrule
\parbox{1.35701cm}{\citep{Chaudhuri:2019}} & \parbox{2.061cm}{Image\\pairs} & \parbox{2.075cm}{Random\\initialization} & \parbox{2.093cm}{Graph\\convolutional\\network} & \parbox{2.081cm}{Metric\\learning\\(supervised)} & \parbox{2cm}{Contrastive} \\\midrule
\parbox{1.35701cm}{\citep{Cao:2019}} & \parbox{2.061cm}{Image\\triplets} & \parbox{2.075cm}{Pre-trained\\network weights} & \parbox{2.093cm}{Convolutional\\neural network} & \parbox{2.081cm}{Metric\\learning\\(supervised)} & \parbox{2cm}{Triplet} 
\\\bottomrule
\end{tabular}}{}
\end{table}
One possible way to overcome this limitation can be an identification of image triplets through visual interpretation instead of defining triplets based on the class labels. Tabular overview of the recent DL based CBIR systems in RS is presented in \Cref{tab2}.

\section{Scalable RS CBIR based on Deep Hashing}
\label{sec:deep_hashing_section}

Due to the significant growth of RS image archives, an image search and retrieval through linear scan (which exhaustively compares the query image with each image in the archive) is computationally expensive and thus impractical. This problem is also known as large-scale CBIR problem. In large-scale CBIR, the storage of the data is also challenging as RS image contents are often represented in high-dimensional features. Accordingly, in addition to the scalability problem, the storage of the image features (descriptors) also becomes a critical bottleneck. To address these problems, approximate nearest neighbor (ANN) search has attracted extensive research attention in RS. In particular, hashing based ANN search schemes have become a cutting-edge research topic for large-scale RS image retrieval due to their high efficiency in both storage cost and search /retrieval speed. Hashing methods encode high-dimensional image descriptors into a low-dimensional Hamming space where the image descriptors are represented by binary hash codes. By this way, the (approximate) nearest neighbors among the images can be efficiently identified based on the the Hamming distance with simple bit-wise operations. In addition, the binary codes can significantly reduce the amount of memory required for storing the content of images. Traditional hashing-based RS CBIR systems initially extract hand-crafted image descriptors and then utilize hash functions that map the original high-dimensional representations into low-dimensional binary codes, such that the similarity to the original space can be well preserved \citep{Demir:2016}, \citep{Li:2017}, \citep{Reato:2019}, \citep{Beltran:2020}. Thus, descriptor extraction and binary code generation are applied independently from each other, resulting in sub-optimal hash codes. Success of DNNs in image feature learning has inspired research on developing DL based hashing methods (i.e., deep hashing methods).

\begin{table}[t]
    \centering
    \footnotesize
    \renewcommand{\arraystretch}{0.5}
    \caption{Main characteristics of the state-of-the-art deep hashing based CBIR systems in RS.\label{tb:loss_cat}}{
    \begin{tabular}{@{}lccc@{}}
    \toprule
    \parbox{1.35015cm}{\raggedright\textbf{{Reference}}} & \parbox{6.081cm}{\raggedright \textbf{Loss Functions}} & \parbox{2.398cm}{\raggedright \textbf{Learning\\Type}} & \parbox{1.114cm}{\raggedright \textbf{Hash Layer}} \\
    \midrule
    \parbox{1.35015cm}{\raggedright \citep{li2017large}} & \parbox{6.081cm}{\raggedright Contrastive, Quantization} & \parbox{2.398cm}{\raggedright supervised} & \parbox{1.114cm}{\raggedright  linear} \\
    \midrule
    \parbox{1.35015cm}{\raggedright \citep{li2018learning}} & \parbox{6.081cm}{\raggedright Contrastive, Quantization} & \parbox{2.398cm}{\raggedright supervised} & \parbox{1.114cm}{\raggedright  linear} \\
    \midrule
    \parbox{1.35015cm}{\raggedright \citep{roy2019metric}} & \parbox{6.081cm}{\raggedright Triplet, Bit balance, Quantization} & \parbox{2.398cm}{\raggedright supervised} & \parbox{1.114cm}{\raggedright sigmoid} \\
    \midrule
    \parbox{1.35015cm}{\raggedright \citep{song2019deep}} & \parbox{6.081cm}{\raggedright Contrastive, Quantization, Cross-entropy} & \parbox{2.398cm}{\raggedright supervised} & \parbox{1.114cm}{\raggedright  linear} \\
    \midrule
    \parbox{1.35015cm}{\raggedright \citep{tang2019remote}} & \parbox{6.081cm}{\raggedright Cross-entropy, Contrastive, Reconstruction, Quantization, Bit balance} & \parbox{2.398cm}{\raggedright semi-supervised} & \parbox{1.114cm}{\raggedright  linear} \\
    \midrule
    \parbox{1.35015cm}{\raggedright \citep{liu2019adversarial}} & \parbox{6.081cm}{\raggedright Adversarial, Quantization, Contrastive, Cross-entropy}& \parbox{2.398cm}{\raggedright supervised} & \parbox{1.114cm}{\raggedright sigmoid} \\
    \bottomrule
    \end{tabular}
    }{}
\end{table}

Recently, several deep hashing based CBIR systems that simultaneously learn image representations and hash functions based on the suitable loss functions are introduced in RS (see \Cref{tb:loss_cat}). As an example, in \citep{li2017large} a supervised deep hashing neural network (DHNN) that learns deep features and binary hash codes by using the contrastive and quantization loss functions in an end-to-end manner is introduced. The contrastive loss function can also be considered as the binary cross-entropy loss function, which is optimized to classify whether an input image pair is similar or not. One advantage of the contrastive loss function is its capability of similarity learning, where similar images can be grouped together, while moving away dissimilar images from each other in the feature space. Due to the ill-posed gradient problem, the standard back-propagation of DL models to directly optimize hash codes is not feasible. The use of the quantization loss mitigates the performance degradation of the generated hash codes through the binarization on the CNN outputs. In \citep{li2018learning} the quantization and contrastive loss functions are combined in the framework of the source-invariant deep hashing CNNs for learning a cross-modality hashing system. Without introducing a margin threshold between the similar and dissimilar images, a limited image retrieval performance can be achieved based on the contrastive loss function. To address this issue, a metric-learning based supervised deep hashing network (MiLaN) is recently introduced in \citep{roy2019metric}. MiLaN is trained by using three different loss functions: 1) the triplet loss function for learning a metric space (where semantically similar images are close to each other and dissimilar images are separated); 2) the bit balance loss function (which aims at forcing the hash codes to have a balanced number of binary values); and 3) the quantization loss function. The bit balance loss function makes each bit of hash codes to have a $50\%$ chance of being activated, and different bits to be independent from each other. As noted in \citep{roy2019metric}, the learned hash codes based on the considered loss functions can efficiently characterize the complex semantics in RS images. A supervised deep hashing CNN (DHCNN) is proposed in \citep{song2019deep} in order to retrieve the semantically similar images in an end-to-end manner. In detail, DHCNN utilizes the joint loss function composed of: 1) the contrastive loss function; 2) the cross-entropy loss function (which aims at increasing the class discrimination capability of hash codes); and 3) the quantization loss. In order to predict the classes based on the hash codes, a FC layer is connected to the hash layer in DHCNN. As mentioned above, one disadvantage of the cross-entropy loss function is its deficiency to define a metric space, where similar images are clustered together. To address this issue, the contrastive loss function is jointly optimized with the cross-entropy loss function in DHCNN. A semi-supervised deep hashing method based on the adversarial autoencoder network (SSHAAE) is proposed in \citep{tang2019remote} for RS CBIR problems. In order to generate the discriminative and similarity preserved hash codes with low quantization errors, SSHAAE exploits the joint loss function composed of: 1) the cross-entropy loss function; 2) a reconstruction loss function; 3) the contrastive loss function; 4) the bit balance loss function; and 5) the quantization loss function. By minimizing the reconstruction loss function, the label vectors and hash codes can be obtained as the latent outputs of the AEs. 
A supervised deep hashing method based on a generative adversarial network (GAN) is proposed in \citep{liu2019adversarial}. For the generator of the GAN, this method introduces a joint loss function that composed of: 1) the cross-entropy loss function; 2) the contrastive loss function; and 3) the quantization loss function. For the discriminator of the GAN, the sigmoid function is used for the classification of the generated hash codes as true codes. This allows to restrict the learned hash codes following the uniform binary distribution. Thus, the bit balance capability of hash codes can be achieved. It is worth noting that the above-mentioned supervised deep hashing methods preserve the discrimination capability and the semantic similarity of the hash codes in the Hamming space by using annotated training images. 
\begin{table}[t]
    \centering
    \footnotesize
    \renewcommand{\arraystretch}{0.5}
    \caption{Comparison of the DL loss functions considered within the deep hashing based RS CBIR systems. Different marks are provided: "\xmark" (no) or "\cmark"(yes).\label{tb:char_loss_func}}{\scalebox{1.0}{
    \begin{tabular}{@{}lccccc}
    \toprule
        \parbox{2.0809cm}{\textbf{Loss\\Function}} & \parbox{1.511cm}{\centering\textbf{Similarity\\Learning\\Capability}} & \parbox{2.6451cm}{\centering\textbf{Mini-batch\\Sampling\\Requirement}} & \parbox{1.453cm}{\centering\textbf{Bit\\Balance\\Capability}} & \parbox{2.081cm}{\centering\textbf{Binarization\\Capability}} & \parbox{1.557cm}{\centering\textbf{Annotated\\Image\\Requirement}}\\ \midrule
        \parbox{2.0809cm}{Contrastive} & \cmark & \parbox{2.6451cm}{\centering Image pairs} & \xmark & \xmark & \cmark\\ \midrule
        \parbox{2.0809cm}{Triplet} & \cmark & \parbox{2.6451cm}{\centering Image triplets} & \xmark & \xmark & \cmark\\ \midrule
        \parbox{2.0809cm}{Adversarial} & \xmark & \parbox{2.6451cm}{\centering Random selection} & \cmark & \xmark & \xmark\\ \midrule
        \parbox{2.0809cm}{Reconstruction} & \xmark & \parbox{2.6451cm}{\centering Random selection} & \xmark & \xmark & \xmark \\ \midrule
        \parbox{2.0809cm}{Cross-entropy} & \xmark & \parbox{2.6451cm}{\centering Random selection} & \xmark & \xmark & \cmark \\ \midrule
        \parbox{2.0809cm}{Bit balance} & \xmark & \parbox{2.6451cm}{\centering Random selection} & \cmark & \xmark & \xmark \\ \midrule
        \parbox{2.0809cm}{Quantization} & \xmark & \parbox{2.6451cm}{\centering Random selection} & \xmark & \cmark & \xmark\\
    \bottomrule
    \end{tabular}}}{}
\end{table}

In \Cref{tb:char_loss_func}, we analyze and compare all the above-mentioned loss functions based on their: i) capability on similarity learning, ii) requirement on the mini-batch sampling; iii) capability of assessing the bit balance issues; iv) capability of binarization of the image descriptors; and v) requirement on the annotated images. For instance, the contrastive and triplet loss functions have the capabilities to learn the relationship among the images in the feature space, where the semantic similarity of hash codes can be preserved. Regarding to the requirement of mini-batch sampling, pairs of images should be sampled for the contrastive loss function, image triplets should be constructed for the triplet loss function. The bit balance and adversarial loss functions are exploited for learning the hash codes with the uniform binary distribution. It is worth noting that an adversarial loss function can be also exploited for other purposes, such as for image augmentation problems to avoid overfitting \citep{cao2018hashgan}. The quantization loss function enforces the produced low-dimensional features by the DNN models to approximate the binary hash codes. With regard to the requirement on image annotations, the contrastive and triplet loss functions require the semantic labels to construct the relationships among the images.

\section{Discussion and Conclusion}
\label{chp:conclusion}

In this chapter, we presented a literature survey on the most recent CBIR systems for efficient and accurate search and retrieval of RS images from massive archives. We focused our attention on the DL based CBIR systems in RS. We initially analyzed the recent DL based CBIR systems based on: i) the strategies considered for the mini-batch sampling; ii) the approaches used for the initialization of the parameters of the considered DNN models; iii) the type of the considered DNNs; and iv) the strategies used for image representation learning. Then, the most recent methodological developments in RS related to scalable image search and retrieval were discussed. In particular, we reviewed the deep hashing based CBIR systems and analyzed the loss functions considered within these systems based on their: i) capability of similarity learning, ii) requirement on the mini-batch sampling; iii) capability of assessing the bit balance issues; iv) capability of binarization; and v) requirement on the annotated images. Analysis of the loss functions under these factors provides a guideline to select the most appropriate loss function for large-scale RS CBIR problems.

It is worth emphasizing that developing accurate and scalable CBIR systems is becoming more and more important due to the increased number of images in the RS data archives. In this context, the CBIR systems discussed in this chapter are very promising. Despite the promising developments discussed in this chapter (e.g., metric learning, local feature aggregation and graph learning), it is still necessary to develop more advanced CBIR systems. For example, most of the systems are based on the direct use of the CNNs for the retrieval tasks, whereas the adapted CNNs are mainly designed for learning a classification problem and thus model the discrimination of pre-defined classes. Thus, the image descriptors obtained through these networks can not learn a image feature space that is directly optimized for the retrieval problems. Siamese and triplet networks are defined in the context of metric learning in RS to address this problem. However, the image similarity information to train these networks is still provided based on the pre-defined classes, preventing to achieve retrieval specific image descriptors. Thus, CBIR systems that can efficiently learn image features optimized for retrieval problems are needed. Furthermore, the existing supervised DL based CBIR systems require a balanced and complete training set with annotated image pairs or triplets, which is difficult to collect in RS. Learning an accurate CBIR model from imbalanced and incomplete training data is very crucial and thus there is a need for developing systems addressing this problem for operational CBIR applications. Furthermore, the availability of an increased number of multi-source RS images (multispectral, hyperspectral and SAR) associated to the same geographical area motivates the need for effective CBIR systems, which can extract and exploit multi-source image descriptors to achieve rich characterization of RS images (and thus to improve image retrieval performance). However, multi-source RS CBIR has not been explored yet (i.e., all the deep hashing based CBIR systems are defined for images acquired by single sensors). Thus, it is necessary to study CBIR systems that can mitigate the aforementioned problems.

\bibliographystyle{IEEEtran}
\bibliography{defs,refs}%

\begin{thebibliography}{10}
\providecommand{\url}[1]{#1}
\csname url@samestyle\endcsname
\providecommand{\newblock}{\relax}
\providecommand{\bibinfo}[2]{#2}
\providecommand{\BIBentrySTDinterwordspacing}{\spaceskip=0pt\relax}
\providecommand{\BIBentryALTinterwordstretchfactor}{4}
\providecommand{\BIBentryALTinterwordspacing}{\spaceskip=\fontdimen2\font plus
\BIBentryALTinterwordstretchfactor\fontdimen3\font minus
  \fontdimen4\font\relax}
\providecommand{\BIBforeignlanguage}[2]{{%
\expandafter\ifx\csname l@#1\endcsname\relax
\typeout{** WARNING: IEEEtran.bst: No hyphenation pattern has been}%
\typeout{** loaded for the language `#1'. Using the pattern for}%
\typeout{** the default language instead.}%
\else
\language=\csname l@#1\endcsname
\fi
#2}}
\providecommand{\BIBdecl}{\relax}
\BIBdecl

\bibitem{Yang:2013}
Y.~{Yang} and S.~{Newsam}, ``Geographic image retrieval using local invariant
  features,'' \emph{IEEE Transactions on Geoscience and Remote Sensing},
  vol.~51, no.~2, pp. 818--832, February 2013.

\bibitem{Aptoula:2014}
E.~{Aptoula}, ``Remote sensing image retrieval with global morphological
  texture descriptors,'' \emph{IEEE Transactions on Geoscience and Remote
  Sensing}, vol.~52, no.~5, pp. 3023--3034, May 2014.

\bibitem{Tekeste:20118}
I.~{Tekeste} and B.~{Demir}, ``Advanced local binary patterns for remote
  sensing image retrieval,'' in \emph{IEEE International Geoscience and Remote
  Sensing Symposium}, July 2018, pp. 6855--6858.

\bibitem{Dai:2018}
O.~E. {Dai}, B.~{Demir}, B.~{Sankur}, and L.~{Bruzzone}, ``A novel system for
  content-based retrieval of single and multi-label high-dimensional remote
  sensing images,'' \emph{IEEE Journal of Selected Topics in Applied Earth
  Observations and Remote Sensing}, vol.~11, no.~7, pp. 2473--2490, July 2018.

\bibitem{Li:2007}
Y.~{Li} and T.~R. {Bretschneider}, ``Semantic-sensitive satellite image
  retrieval,'' \emph{IEEE Transactions on Geoscience and Remote Sensing},
  vol.~45, no.~4, pp. 853--860, April 2007.

\bibitem{Chaudhuri:2016}
B.~{Chaudhuri}, B.~{Demir}, L.~{Bruzzone}, and S.~{Chaudhuri}, ``Region-based
  retrieval of remote sensing images using an unsupervised graph-theoretic
  approach,'' \emph{IEEE Geoscience and Remote Sensing Letters}, vol.~13,
  no.~7, pp. 987--991, July 2016.

\bibitem{Chaudhuri:2018}
B.~{Chaudhuri}, B.~{Demir}, S.~{Chaudhuri}, and L.~{Bruzzone}, ``Multilabel
  remote sensing image retrieval using a semisupervised graph-theoretic
  method,'' \emph{IEEE Transactions on Geoscience and Remote Sensing}, vol.~56,
  no.~2, pp. 1144--1158, February 2018.

\bibitem{Demir:2016}
B.~{Demir} and L.~{Bruzzone}, ``Hashing-based scalable remote sensing image
  search and retrieval in large archives,'' \emph{IEEE Transactions on
  Geoscience and Remote Sensing}, vol.~54, no.~2, pp. 892--904, February 2016.

\bibitem{Li:2017}
P.~{Li} and P.~{Ren}, ``Partial randomness hashing for large-scale remote
  sensing image retrieval,'' \emph{IEEE Geoscience and Remote Sensing Letters},
  vol.~14, no.~3, pp. 464--468, March 2017.

\bibitem{Reato:2019}
T.~{Reato}, B.~{Demir}, and L.~{Bruzzone}, ``An unsupervised multicode hashing
  method for accurate and scalable remote sensing image retrieval,'' \emph{IEEE
  Geoscience and Remote Sensing Letters}, vol.~16, no.~2, pp. 276--280, October
  2019.

\bibitem{Demir:2015}
B.~{Demir} and L.~{Bruzzone}, ``A novel active learning method in relevance
  feedback for content-based remote sensing image retrieval,'' \emph{IEEE
  Transactions on Geoscience and Remote Sensing}, vol.~53, no.~5, pp.
  2323--2334, May 2015.

\bibitem{Wang:2016}
Y.~{Wang}, L.~{Zhang}, X.~{Tong}, L.~{Zhang}, Z.~{Zhang}, H.~{Liu}, X.~{Xing},
  and P.~T. {Mathiopoulos}, ``A three-layered graph-based learning approach for
  remote sensing image retrieval,'' \emph{IEEE Transactions on Geoscience and
  Remote Sensing}, vol.~54, no.~10, pp. 6020--6034, October 2016.

\bibitem{Zhou:2015}
W.~Zhou, Z.~Shao, C.~Diao, and Q.~Cheng, ``High-resolution remote-sensing
  imagery retrieval using sparse features by auto-encoder,'' \emph{Remote
  Sensing Letters}, vol.~6, no.~10, pp. 775--783, October 2015.

\bibitem{Li:2016}
Y.~Li, Y.~Zhang, C.~Tao, and H.~Zhu, ``Content-based high-resolution remote
  sensing image retrieval via unsupervised feature learning and collaborative
  affinity metric fusion,'' \emph{Remote Sensing}, vol.~8, no.~9, p. 709,
  August 2016.

\bibitem{Xu:2018}
X.~Tang, X.~Zhang, F.~Liu, and L.~Jiao, ``Unsupervised deep feature learning
  for remote sensing image retrieval,'' \emph{Remote Sensing}, vol.~10, no.~8,
  p. 1243, August 2018.

\bibitem{Hu:2016}
F.~{Hu}, X.~{Tong}, G.~{Xia}, and L.~{Zhang}, ``Delving into deep
  representations for remote sensing image retrieval,'' in \emph{IEEE
  International Conference on Signal Processing}, November 2016, pp. 198--203.

\bibitem{Zhou:2017}
W.~Zhou, S.~Newsam, C.~Li, and Z.~Shao, ``Learning low dimensional
  convolutional neural networks for high-resolution remote sensing image
  retrieval,'' \emph{Remote Sensing}, vol.~9, no.~5, p. 489, May 2017.

\bibitem{Ye:2018}
F.~{Ye}, H.~{Xiao}, X.~{Zhao}, M.~{Dong}, W.~{Luo}, and W.~{Min}, ``Remote
  sensing image retrieval using convolutional neural network features and
  weighted distance,'' \emph{IEEE Geoscience and Remote Sensing Letters},
  vol.~15, no.~10, pp. 1535--1539, October 2018.

\bibitem{Boualleg:2018}
Y.~{Boualleg} and M.~{Farah}, ``Enhanced interactive remote sensing image
  retrieval with scene classification convolutional neural networks model,'' in
  \emph{IEEE International Geoscience and Remote Sensing Symposium}, July 2018,
  pp. 4748--4751.

\bibitem{Imbriaco:2019}
R.~Imbriaco, C.~Sebastian, E.~Bondarev, and P.~H.~N. de~With, ``Aggregated deep
  local features for remote sensing image retrieval,'' \emph{Remote Sensing},
  vol.~11, p. 493, February 2019.

\bibitem{Chaudhuri:2019}
U.~Chaudhuri, B.~Banerjee, and A.~Bhattacharya, ``Siamese graph convolutional
  network for content based remote sensing image retrieval,'' \emph{Computer
  Vision and Image Understanding}, vol. 184, pp. 22--30, July 2019.

\bibitem{Cao:2019}
R.~Cao, Q.~Zhang, J.~Zhu, Q.~Li, Q.~Li, B.~Liu, and G.~Qiu, ``Enhancing remote
  sensing image retrieval using a triplet deep metric learning network,''
  \emph{International Journal of Remote Sensing}, vol.~41, no.~2, pp. 740--751,
  January 2020.

\bibitem{Beltran:2020}
R.~{Fernandez-Beltran}, B.~{Demir}, F.~{Pla}, and A.~{Plaza}, ``Unsupervised
  remote sensing image retrieval using probabilistic latent semantic hashing,''
  \emph{IEEE Geoscience and Remote Sensing Letters}, February 2020.

\bibitem{li2017large}
Y.~{Li}, Y.~{Zhang}, X.~{Huang}, H.~{Zhu}, and J.~{Ma}, ``Large-scale remote
  sensing image retrieval by deep hashing neural networks,'' \emph{IEEE
  Transactions on Geoscience and Remote Sensing}, vol.~56, no.~2, pp. 950--965,
  February 2018.

\bibitem{li2018learning}
Y.~{Li}, Y.~{Zhang}, X.~{Huang}, and J.~{Ma}, ``Learning source-invariant deep
  hashing convolutional neural networks for cross-source remote sensing image
  retrieval,'' \emph{IEEE Transactions on Geoscience and Remote Sensing},
  vol.~56, no.~11, pp. 6521--6536, June 2018.

\bibitem{roy2019metric}
S.~{Roy}, E.~{Sangineto}, B.~{Demir}, and N.~{Sebe}, ``Metric-learning-based
  deep hashing network for content-based retrieval of remote sensing images,''
  \emph{IEEE Geoscience and Remote Sensing Letters}, February 2020.

\bibitem{song2019deep}
W.~{Song}, S.~{Li}, and J.~{Benediktsson}, ``Deep hashing learning for visual
  and semantic retrieval of remote sensing images,'' \emph{arXiv preprint
  arXiv:1909.04614}, 2019.

\bibitem{tang2019remote}
X.~{Tang}, C.~{Liu}, X.~{Zhang}, J.~{Ma}, C.~{Jiao}, and L.~{Jiao}, ``Remote
  sensing image retrieval based on semi-supervised deep hashing learning,'' in
  \emph{IEEE International Geoscience and Remote Sensing Symposium}, July 2019,
  pp. 879--882.

\bibitem{liu2019adversarial}
C.~{Liu}, J.~{Ma}, X.~{Tang}, X.~{Zhang}, and L.~{Jiao}, ``Adversarial
  hash-code learning for remote sensing image retrieval,'' in \emph{IEEE
  International Geoscience and Remote Sensing Symposium}, July 2019, pp.
  4324--4327.

\bibitem{cao2018hashgan}
Y.~{Cao}, B.~{Liu}, M.~{Long}, and J.~{Wang}, ``Hashgan: Deep learning to hash
  with pair conditional wasserstein gan,'' in \emph{IEEE Conference on Computer
  Vision and Pattern Recognition}, June 2018, pp. 1287--1296.

\end{thebibliography}

\end{document}